\documentclass[review]{elsarticle}
\usepackage{lineno,hyperref}
\usepackage{amsmath,graphicx}
\usepackage{amssymb}
\usepackage{epstopdf}
\usepackage{algorithm}
\usepackage{algpseudocode}
\usepackage{tabularx}
\usepackage{tabulary}
\usepackage{caption}
\usepackage{amsmath,bm}
\usepackage{colortbl,booktabs}
\usepackage{subfig}
\modulolinenumbers[5]
\journal{Journal of \LaTeX\ Templates}
\begin{document}
\begin{frontmatter}

\title{Bias-Compensated Normalized Maximum Correntropy Criterion Algorithm for System Identification with Noisy Input}

\author{Elsevier\fnref{myfootnote}}
\address{Radarweg 29, Amsterdam}
\fntext[myfootnote]{Since 1880.}

\author[mymainaddress]{Wentao Ma\corref{mycorrespondingauthor}}
\cortext[mycorrespondingauthor]{Corresponding author}
\ead{mawt@xaut.edu.cn}

\author[mymainaddress]{Dongqiao Zheng}
\author[mysecondaryaddress]{Yuanhao Li}
\author[mymainaddress]{Zhiyu Zhang}
\author[mysecondaryaddress]{Badong Chen}

\address[mymainaddress]{School of Automation and Information Engineering, Xi'an University of Technology, China}
\address[mysecondaryaddress]{School of Electronic and Information Engineering, Xi'an Jiaotong University, China}

\begin{abstract}
This paper proposed a bias-compensated normalized maximum correntropy criterion (BCNMCC) algorithm charactered by its low steady-state misalignment for system identification with noisy input in an impulsive output noise environment. The normalized maximum correntropy criterion (NMCC) is derived from a correntropy based cost function, which is rather robust with respect to impulsive noises. To deal with the noisy input, we introduce a bias-compensated vector (BCV) to the NMCC algorithm, and then an unbiasedness criterion and some reasonable assumptions are used to compute the BCV. Taking advantage of the BCV, the bias caused by the input noise can be effectively suppressed. System identification simulation results demonstrate that the proposed BCNMCC algorithm can outperform other related algorithms with noisy input especially in an impulsive output noise environment.
\end{abstract}

\end{frontmatter}
\linenumbers

\section{Introduction}
Recently, bias-compensated adaptive filtering algorithms (BCAFAs) \cite{kang2013bias,zhao2016bias,zheng2016bias,zheng2017bias} based on the unbiasedness criterion (UC) are paid attention to in several signal processing applications in a noisy input case, that is, a bias-compensated vector (BCV) is introduced to reduce the bias caused by the input noise. In particular, the bias-compensated normalized least mean square (BCNLMS) algorithm is popular due to its simplicity and effectiveness \cite{kang2013bias}. In \cite{zhao2016bias}, the bias-compensated affine projection algorithm (APA) was developed to reduce the performance degradation caused by highly correlated input. The bias-compensated normalized subband adaptive filter algorithm was proposed in \cite{zheng2016bias}, which has a better performance and does not require input-output variance ration advance. The bias-compensated normalized least mean fourth (NLMF) was presented in \cite{zheng2017bias}, which can offer a faster convergence rate and a lower steady-state misalignment in a certain case. Furthermore, the BCNLMS with $L_1$-norm was proposed in \cite{yoo2015improved} to address the noisy input problem in sparse system identification. At present in \cite{jung2017stabilization,wang2018bias}, the convergence analysis of the BCNLMS has been performed based on some suitable assumptions. All the above BCAFAs have been successfully utilized to solve the noisy input problem in different applications. However, they are sensitive to output noise with impulsive characters.

In order to improve the robustness with respect to output noise, some improved adaptive filtering algorithms (AFAs) have been proposed to eliminate the bad influence of the output noise in different literatures \cite{arikan1994adaptive,zou2000recursive,singh2009using,haddad2016unified,zayyani2014continuous}. Particularly, different kinds of AFAs based on maximum correntropy criterion (MCC) were developed \cite{liu2007correntropy,chen2012maximum}, such as sparse MCC \cite{ma2015maximum}, diffusion MCC \cite{ma2016diffusion}, kernel MCC \cite{wu2015kernel} and so on. Although the MCC based AFAs can improve the robustness in non-Gaussian signal processing, a noisy input case is however not considered in these solutions and thus they are sensitive to the scaling of input signals.

Considering the drawbacks of the existing BCAFAs and the MCC based AFAs, we take advantage of the UC and robust property of the MCC to develop a novel bias-compensated normalized MCC (BCNMCC) algorithm in this study in order to eliminate the influence of the input noise and the impulsive output noise.

The rest of this paper is structured as follows. In Section 2, the NMCC algorithm is briefly reviewed. In Section 3, we develop the bias-compensated NMCC algorithm. In Section 4, simulation experiments are conducted to demonstrate the performance of the new method. Finally, the paper is concluded in Section 5.

\section{Review of the NMCC algorithm}
For an adaptive filter under a common system identification (SI) framework, the desired signal is generally denoted by
\begin{equation}
d(i)={{\mathbf{u}}^{T}}(i){{\mathbf{w}}^{o}}+v(i)
\end{equation}
where ${{\mathbf{w}}^{o}}={{[w_{1}^{o},w_{2}^{o},...w_{L}^{o}]}^{T}}$ denotes an unknown system parameter vector with $L$-tap to be estimated, and the perturbation signal $v(i)$ is the output noise at time index $i$. $\mathbf{u}(i)={{[{{u}_{1}}(i),{{u}_{2}}(i),...,{{u}_{L}}(i)]}^{T}}$ denotes the input vector. In \cite{singh2009using}, the update of the MCC based AFA is given by
\begin{equation}
\mathbf{w}(i+1)=\mathbf{w}(i)+\mu \exp (-\frac{{{e}^{2}}(i)}{2{{\sigma }^{2}}})e(i)\mathbf{u}(i)
\end{equation}
where $\mathbf{w}(i)={{[{{w}_{1}}(i),{{w}_{2}}(i),...{{w}_{L}}(i)]}^{T}}$ denotes the tap-coefficients vector of an adaptive filter which is employed to find an estimate of ${{w}^{o}}$ from the observed input-output data. $\mu$ is the step size and $\sigma$ denotes the kernel bandwidth, which is a positive parameter that induces a trade-off between convergence speed and steady-state accuracy. $e(i)=d(i)-{{\mathbf{u}}^{T}}(i)\mathbf{w}(i)$ denotes the ${{i}^{th}}$ instantaneous error. Considering the MCC and the idea of the normalized least mean square, the normalized MCC (NMCC) updating equation can be represented as
\begin{equation}
\mathbf{w}(i+1)=\mathbf{w}(i)+\mu \exp (-\frac{{{e}^{2}}(i)}{2{{\sigma }^{2}}})\frac{e(i)\mathbf{u}(i)}{{{\mathbf{u}}^{T}}(i)\mathbf{u}(i)+\varepsilon }
\end{equation}
where $\varepsilon$ is a regularization parameter.

\section{Bias-compensated NMCC}
In this section, we focus on developing the bias-compensated NMCC algorithm based on UC and the NMCC algorithm in (3) for system identification problem in a noisy input and output environment. Considering the input noise, we define the input vector as
\begin{equation}
\mathbf{\bar{u}}(i)=\mathbf{u}(i)+\mathbf{\eta }(i)
\end{equation}
where $\mathbf{\eta }(i)={{[\eta (i),\eta (i-1),...\eta (i-L+1)]}^{T}}$ is the noise vector, and ${{\mathbf{\eta }}_{l}}(i)(l\in [1,L])$ is with zero-mean Gaussian and variance $\sigma_{in}^{2}$. One can rewrite the filtered output error as
\begin{equation}
\begin{split}
  & \bar{e}(i)=d(i)-{{{\mathbf{\bar{u}}}}^{T}}(i)\mathbf{w}(i) \\
 & =d(i)-{{(\mathbf{u}(i)+\mathbf{\eta }(i))}^{T}}\mathbf{w}(i) \\
 & =\mathbf{u}{{(i)}^{T}}\mathbf{\tilde{w}}(i)+v(i)-\mathbf{\eta }{{(i)}^{T}}\mathbf{w}(i) \\
 & ={{e}_{w}}(i)+v(i)-\mathbf{\eta }{{(i)}^{T}}\mathbf{w}(i) \\
\end{split}
\end{equation}
where ${{e}_{w}}(i)={{\mathbf{u}}^{T}}(i)\mathbf{\tilde{w}}(i)$ is the priori error and the weight-error vector is denoted as $\mathbf{\tilde{w}}(i)={{\mathbf{w}}_{o}}-\mathbf{w}(i)$. To compensate the bias caused by the input noise, we introduce a bias-compensation vector $\mathbf{B}(i)$ into (5), and replace $\mathbf{u}(i)$ and $e(i)$ with $\mathbf{\bar{u}}(i)$ and $\bar{e}(i)$ simultaneously. The equation (5) is improved as
\begin{equation}
\mathbf{w}(i+1)=\mathbf{w}(i)+\mu f(\bar{e}(i))\frac{\bar{e}(i)\mathbf{\bar{u}}(i)}{{{{\mathbf{\bar{u}}}}^{T}}(i)\mathbf{\bar{u}}(i)+\varepsilon }+\mathbf{B}(i)
\end{equation}
where $f(\bar{e}(i))$ is a non-linear function of the estimation error, which is defined as
\begin{equation}
f(\bar{e}(i))=\exp \left( -\frac{{{{\bar{e}}}^{2}}(i)}{2{{\sigma }^{2}}} \right)=\exp \left( -\frac{{{({{e}_{w}}(i)+v(i)-{{\mathbf{\eta }}^{T}}(i)\mathbf{w}(i))}^{2}}}{2{{\sigma }^{2}}} \right)
\end{equation}
Then, combining (6) and the definition of the $\mathbf{\tilde{w}}(i)$, one can obtain
\begin{equation}
\mathbf{\tilde{w}}(i+1)=\mathbf{\tilde{w}}(i)-\mu f(\bar{e}(i))\frac{\bar{e}(i)\mathbf{\bar{u}}(i)}{{{{\mathbf{\bar{u}}}}^{T}}(i)\mathbf{\bar{u}}(i)+\varepsilon }-\mathbf{B}(i)
\end{equation}

Now, we employ the unbiasedness criterion in (9) to get the bias-compensated vector.
\begin{equation}
E(\mathbf{\tilde{w}}(i+1)\left| \mathbf{\bar{u}}(i) \right.)=0 \qquad whenever \qquad E(\mathbf{\tilde{w}}(i)\left| \mathbf{\bar{u}}(i) \right.)=0
\end{equation}

Taking expectation on both sides of (8) with the given $\mathbf{\bar{u}}(i)$ and using criterion (9), one can obtain
\begin{equation}
\begin{split}
  & E\left[ \mathbf{\tilde{w}}(i+1)\left| \mathbf{\bar{u}}(i) \right. \right]= \\
 & E\left[ \mathbf{\tilde{w}}(i)\left| \mathbf{\bar{u}}(i) \right. \right]-\mu E\left[ f(\bar{e}(i))\frac{\bar{e}(i)\mathbf{\bar{u}}(i)}{{{{\mathbf{\bar{u}}}}^{T}}(i)\mathbf{\bar{u}}(i)+\varepsilon }\left| \mathbf{\bar{u}}(i) \right. \right]-E\left[ \mathbf{B}(i)\left| \mathbf{\bar{u}}(i) \right. \right] \\
\end{split}
\end{equation}

According to (9) and (10), the following equation is obtained
\begin{equation}
E\left[ \mathbf{B}(i)\left| \mathbf{\bar{u}}(i) \right. \right]=-\mu E\left[ f(\bar{e}(i))\frac{\bar{e}(i)\mathbf{\bar{u}}(i)}{{{{\mathbf{\bar{u}}}}^{T}}(i)\mathbf{\bar{u}}(i)+\varepsilon }\left| \mathbf{\bar{u}}(i) \right. \right]
\end{equation}

In order to calculate the gradient of the BCNMCC algorithm, the following commonly-used assumptions \cite{shin2004variable,zhang2014transient,chen2014steady} are given:

\textbf{Assumption 1:} The background noise $v(i)$ is zero-mean $\alpha$-stable distribution noise and input noise $\mathbf{\eta }(i)$ is zero-mean white Gaussian noise.

\textbf{Assumption 2:} The signals $v(i)$, $\mathbf{\eta }(i)$, $\mathbf{u}(i)$ and $\mathbf{\tilde{w}}(i)$ are statistically independent.

\textbf{Assumption 3:} The non-linear function of the estimation error $f(v(i))$, $\mathbf{\eta }(i)$ and $\bar{e}(i)$ are statistically independent.

To simplify the following analysis, we take the Taylor expansion of $f(\bar{e}(i))$ with respect to ${{e}_{w}}(i)-{{\mathbf{\eta }}^{T}}(i)\mathbf{w}(i)$ around $v(i)$. Combining (5) one can obtain
\begin{equation}
f(\bar{e}(i))\approx f(v(i))+f'(v(i))[{{e}_{w}}(i)-{{\mathbf{\eta }}^{T}}(i)\mathbf{w}(i)]+o\left[ {{[{{e}_{w}}(i)-{{\mathbf{\eta }}^{T}}(i)\mathbf{w}(i)]}^{\text{2}}} \right]
\end{equation}

From (11), the following approximation can be obtained
\begin{equation}
\begin{split}
  & E\left[ f(\bar{e}(i))\frac{\bar{e}(i)\mathbf{\bar{u}}(i)}{{{{\mathbf{\bar{u}}}}^{T}}(i)\mathbf{\bar{u}}(i)+\varepsilon }\left| \mathbf{\bar{u}}(i) \right. \right] \\
 & \approx E\left[ f(v(i))\frac{\bar{e}(i)\mathbf{\bar{u}}(i)}{{{{\mathbf{\bar{u}}}}^{T}}(i)\mathbf{\bar{u}}(i)+\varepsilon }\left| \mathbf{\bar{u}}(i) \right. \right] \\
 & +E\left[ f'(v(i))[{{e}_{w}}(i)-{{\mathbf{\eta }}^{T}}(i)\mathbf{w}(i)]\frac{\bar{e}(i)\mathbf{\bar{u}}(i)}{{{{\mathbf{\bar{u}}}}^{T}}(i)\mathbf{\bar{u}}(i)+\varepsilon }\left| \mathbf{\bar{u}}(i) \right. \right] \\
 & +E\left[ o\left[ {{[{{e}_{w}}(i)-{{\mathbf{\eta }}^{T}}(i)\mathbf{w}(i)]}^{\text{2}}} \right]\frac{\bar{e}(i)\mathbf{\bar{u}}(i)}{{{{\mathbf{\bar{u}}}}^{T}}(i)\mathbf{\bar{u}}(i)+\varepsilon }\left| \mathbf{\bar{u}}(i) \right. \right]
\end{split}
\end{equation}

In the steady-state, the priori error ${{e}_{w}}(i)$ converges to a small value which is ignorable with respect to the environmental noise when the step size is small \cite{chen2014steady}. Considering the assumptions 1, 2 and 3, the second term of equation (13) becomes
\begin{equation}
\begin{split}
  & E\left[ f'(v(i))[{{e}_{w}}(i)-{{\mathbf{\eta }}^{T}}(i)\mathbf{w}(i)]\frac{\bar{e}(i)\mathbf{\bar{u}}(i)}{{{{\mathbf{\bar{u}}}}^{T}}(i)\mathbf{\bar{u}}(i)+\varepsilon }\left| \mathbf{\bar{u}}(i) \right. \right] \\
 & \approx -E\left[ f'(v(i)){{\mathbf{\eta }}^{T}}(i)\mathbf{w}(i)\frac{\bar{e}(i)\mathbf{\bar{u}}(i)}{{{{\mathbf{\bar{u}}}}^{T}}(i)\mathbf{\bar{u}}(i)+\varepsilon }\left| \mathbf{\bar{u}}(i) \right. \right]\text{=0}
\end{split}
\end{equation}
Similarly the third term of equation (13) is
\begin{equation}
E\left[ o\left[ {{[{{e}_{w}}(i)-{{\mathbf{\eta }}^{T}}(i)\mathbf{w}(i)]}^{\text{2}}} \right]\frac{\bar{e}(i)\mathbf{\bar{u}}(i)}{{{{\mathbf{\bar{u}}}}^{T}}(i)\mathbf{\bar{u}}(i)+\varepsilon }\left| \mathbf{\bar{u}}(i) \right. \right]\text{=0}
\end{equation}
Combining (13), (14) and (15), and using assumption 3, we have
\begin{equation}
\begin{split}
  & E\left[ f(\bar{e}(i))\frac{\bar{e}(i)\mathbf{\bar{u}}(i)}{{{{\mathbf{\bar{u}}}}^{T}}(i)\mathbf{\bar{u}}(i)+\varepsilon }\left| \mathbf{\bar{u}}(i) \right. \right]\approx E\left[ f(v(i))\frac{\bar{e}(i)\mathbf{\bar{u}}(i)}{{{{\mathbf{\bar{u}}}}^{T}}(i)\mathbf{\bar{u}}(i)+\varepsilon }\left| \mathbf{\bar{u}}(i) \right. \right] \\
 & =E\left[ f(v(i))\left| \mathbf{\bar{u}}(i) \right. \right]E\left[ \frac{\bar{e}(i)\mathbf{\bar{u}}(i)}{{{{\mathbf{\bar{u}}}}^{T}}(i)\mathbf{\bar{u}}(i)+\varepsilon }\left| \mathbf{\bar{u}}(i) \right. \right] \\
\end{split}
\end{equation}
Considering the fact that $\bar{e}(i)=e(i)-{{\mathbf{\eta }}^{T}}(i)\mathbf{w}(i)$
\begin{equation}
E\left[ \frac{\bar{e}(i)\mathbf{\bar{u}}(i)}{{{{\mathbf{\bar{u}}}}^{T}}(i)\mathbf{\bar{u}}(i)+\varepsilon }\left| \mathbf{\bar{u}}(n) \right. \right]=\frac{E[\bar{e}(i)\mathbf{\bar{u}}(i)\left| \mathbf{\bar{u}}(i) \right.]}{{{{\mathbf{\bar{u}}}}^{T}}(i)\mathbf{\bar{u}}(i)+\varepsilon }
\end{equation}
\begin{equation}
\begin{split}
  & E\left[ \bar{e}(i)\mathbf{\bar{u}}(i)\left| \mathbf{\bar{u}}(i) \right. \right] \\
 & =E\left[ [e(i)-{{\mathbf{\eta }}^{T}}(i)\mathbf{w}(i)][\mathbf{u}(i)+\mathbf{\eta }(i)]\left| \mathbf{\bar{u}}(i) \right. \right] \\
 & =E\left[ e(i)\mathbf{\bar{u}}(i)\left| \mathbf{\bar{u}}(i) \right. \right]+E\left[ e(i)\mathbf{\eta }(i)\left| \mathbf{\bar{u}}(i) \right. \right] \\
 & -E\left[ {{\mathbf{\eta }}^{T}}(i)\mathbf{w}(i)\mathbf{u}(i)\left| \mathbf{\bar{u}}(i) \right. \right]-E\left[ {{\mathbf{\eta }}^{T}}(i)\mathbf{w}(i)\mathbf{\eta }(i)\left| \mathbf{\bar{u}}(i) \right. \right]
\end{split}
\end{equation}
\begin{equation}
\begin{split}
  & E\left[ e(i)\mathbf{\bar{u}}(i)\left| \mathbf{\bar{u}}(i) \right. \right] \\
 & =E\left[ [v(i)+{{\mathbf{u}}^{T}}(i)\mathbf{\tilde{w}}(i)][\mathbf{u}(i)+\mathbf{\eta }(i)]\left| \mathbf{\bar{u}}(i) \right. \right] \\
 & =E\left[ v(i)\mathbf{u}(i)\left| \mathbf{\bar{u}}(i) \right. \right]+E\left[ v(i)\mathbf{\eta }(i)\left| \mathbf{\bar{u}}(i) \right. \right] \\
 & +E\left[ {{\mathbf{u}}^{T}}(i)\mathbf{\tilde{w}}(i)\mathbf{u}(i)\left| \mathbf{\bar{u}}(i) \right. \right]+E\left[ {{\mathbf{u}}^{T}}(i)\mathbf{\tilde{w}}(i)\mathbf{\eta }(i)\left| \mathbf{\bar{u}}(i) \right. \right] \\
 & =0 \\
\end{split}
\end{equation}
and
\begin{equation}
E[{{\mathbf{\eta }}^{T}}(i)\mathbf{w}(i)\mathbf{u}(i)\left| \mathbf{\bar{u}}(i) \right.]=0
\end{equation}
\begin{equation}
E[{{\mathbf{\eta }}^{T}}(i)\mathbf{w}(i)\mathbf{\eta }(i)\left| \mathbf{\bar{u}}(i) \right.]=\sigma _{in}^{2}E[\mathbf{w}(i)\left| \mathbf{\bar{u}}(i) \right.]
\end{equation}
Combining (17-20) and (21) we can obtain
\begin{equation}
E\left[ \frac{\bar{e}(i)\mathbf{\bar{u}}(i)}{{{{\mathbf{\bar{u}}}}^{T}}(i)\mathbf{\bar{u}}(i)+\varepsilon }\left| \mathbf{\bar{u}}(i) \right. \right]=-E\left[ \frac{\sigma _{in}^{2}\mathbf{w}(i)}{{{{\mathbf{\bar{u}}}}^{T}}(i)\mathbf{\bar{u}}(i)+\varepsilon }\left| \mathbf{\bar{u}}(i) \right. \right]
\end{equation}
Then, we put (16) and (22) into (11)
\begin{equation}
E\left[ \mathbf{B}(i)\left| \mathbf{\bar{u}}(i) \right. \right]=\mu E\left[ \exp \left( -\frac{{{v}^{2}}(i)}{2{{\sigma }^{2}}} \right)\left| \mathbf{\bar{u}}(i) \right. \right]E\left[ \frac{\sigma _{in}^{2}\mathbf{w}(i)}{{{{\mathbf{\bar{u}}}}^{T}}(i)\mathbf{\bar{u}}(i)+\varepsilon }\left| \mathbf{\bar{u}}(i) \right. \right]
\end{equation}
Considering this condition and the stochastic approximation given in \cite{kang2013bias}, we can obtain the bias-compensated vector as
\begin{equation}
\mathbf{B}(i)=\mu \exp \left( -\frac{{{v}^{2}}(i)}{2{{\sigma }^{2}}} \right)\frac{\sigma _{in}^{2}\mathbf{w}(i)}{{{{\mathbf{\bar{u}}}}^{T}}(i)\mathbf{\bar{u}}(i)+\varepsilon }
\end{equation}
By substituting (24) into (6) we have
\begin{equation}
\begin{split}
   \mathbf{w}(i+1)=&\left( 1+\mu \exp \left( -\frac{{{v}^{2}}(i)}{2{{\sigma }^{2}}} \right)\frac{\sigma _{in}^{2}}{{{{\mathbf{\bar{u}}}}^{T}}(i)\mathbf{\bar{u}}(i)+\varepsilon } \right)\mathbf{w}(i) \\
 & +\mu \exp \left( -\frac{{{{\bar{e}}}^{2}}(i)}{2{{\sigma }^{2}}} \right)\frac{\bar{e}(i)\mathbf{\bar{u}}(i)}{{{{\mathbf{\bar{u}}}}^{T}}(i)\mathbf{\bar{u}}(i)+\varepsilon }
\end{split}
\end{equation}

Generally, the input noise variance is unknown in practice and it is supposed to be estimated suitably. In recent years, several estimation methods have been reported in \cite{jo2005consistent,minkoff2001comment}. In this paper, we utilize the estimation method proposed in \cite{jo2005consistent} to estimate the input noise variance. The estimate equation is given as
\begin{equation}
\sigma _{in}^{\text{2}}(i)=\frac{\sigma _{{\bar{e}}}^{2}(i)}{L\sigma _{\mathbf{w}}^{2}(i)+\kappa +\frac{\sigma _{{\bar{e}}}^{2}(i)L}{{{{\bar{u}}}^{T}}(i)\bar{u}(i)}}
\end{equation}
where $\kappa$ is input-output noise ratio, and it is assumed to be available.
\begin{equation}
\sigma _{{\bar{e}}}^{2}(i)=a\sigma _{{\bar{e}}}^{2}(i-1)+(1-a){{\bar{e}}^{2}}(i)
\end{equation}
\begin{equation}
\sigma _{\mathbf{w}}^{2}(i)=a\sigma _{\mathbf{w}}^{2}(i-1)+(1-a)\frac{1}{L}{{\mathbf{w}}^{T}}(i)\mathbf{w}(i)
\end{equation}
where the parameter $a$ is a forgetting factor.

\section{Simulation results}
In this section, we perform simulations on SI to verify the performance of the proposed BCNMCC algorithm compared to several existing algorithms including LMS, NLMS, and BCNLMS with noisy input in non-Gaussian output noise environments. The input signal is Gaussian with mean 1.0 and unit-variance, and the input noise is assumed to be a white zero-mean Gaussian random sequence. We consider that the output noise is generated by $\alpha$-stable distribution to show the robustness of the proposed algorithm. The characteristic function of the $\alpha$-stable distribution is defined by
\begin{equation}
f(t)=\exp \{j\theta t-\gamma |t{|}^{\alpha }[1+j\beta sgn(t)S(t,\alpha )]\}
\end{equation}
in which
\begin{equation}
S(t,\alpha )=\left\{ \begin{split}
  & \tan \frac{\alpha \pi }{2} \qquad if \quad \alpha \ne 1 \\
 & \frac{2}{\pi }\log |t| \qquad if \quad \alpha =1 \\
\end{split} \right.
\end{equation}
where $\alpha \in (0,2]$ is the characteristic factor, $\text{-}\infty <\theta <+\infty $ is the location parameter, $\beta \in [-1,1]$ is the symmetry parameter, and $\gamma >0$ is the dispersion parameter. The parameters vector of the characteristic function is defined as ${{V}_{\alpha -stable}}(\alpha ,\beta ,\gamma ,\theta )$. 200 Monte-Carlo trials are conducted to obtain the MSD which is defined by
\begin{equation}
MSD=10{{\log }_{10}}\left( E\left( \frac{||{{\mathbf{w}}^{o}}-\mathbf{w}(i)|{{|}^{2}}}{||{{\mathbf{w}}^{o}}|{{|}^{2}}} \right) \right)
\end{equation}

In order to investigate the convergence performance of the proposed BCNMCC, we elaborately select a special system defined by an optimum weight vector ${{\mathbf{w}}^{o}}\text{=}{{\left[ -0.3,-0.9,0.8,-0.7,0.6 \right]}^{T}}$. The parameter settings in the following simulations are abide by to achieve optimal performance for each algorithm. The kernel bandwidth $\sigma$ is 4. The input noise variance ${{\delta }^{2}}=0.25$, and the output noise parameter vector is set as ${{V}_{\alpha -stable}}(1.3,0,0.2,0)$ without special instructions. The input-output noise ratio $\kappa =\text{5}$, and the parameter $\varepsilon =\text{0}\text{.001}$.

In the first example, we examine the convergence performance of different algorithms in terms of MSD with time-varying non-Gaussian impulsive noise generated by different $\alpha$. For the following two stages, the parameter $\alpha$ is set at 1.8 and1.3 respectively, which means the impulsive is increasing. In order to keep fairness, we set a same initial condition for both stages. The convergence curves are given in Fig.1. The step sizes for all algorithms are selected to keep the same initial convergence speed showed in the legend of the Fig.1. One can clearly see that all algorithms can converge while the BCNLMS and BCNMCC have lower steady-state error compared with NLMS and NMCC due to the advantage of the bias-compensated term. In the second stage ($\alpha =\text{1}\text{.3}$), we observe that the LMS and BCNLMS do not converge while the MCC and BCNMCC still converge. Further, the BCNMCC has higher steady state accuracy than MCC because of the introduced bias-compensated term.
\begin{figure}[!t]
\centering
\includegraphics[height=7cm,width=9.5cm]{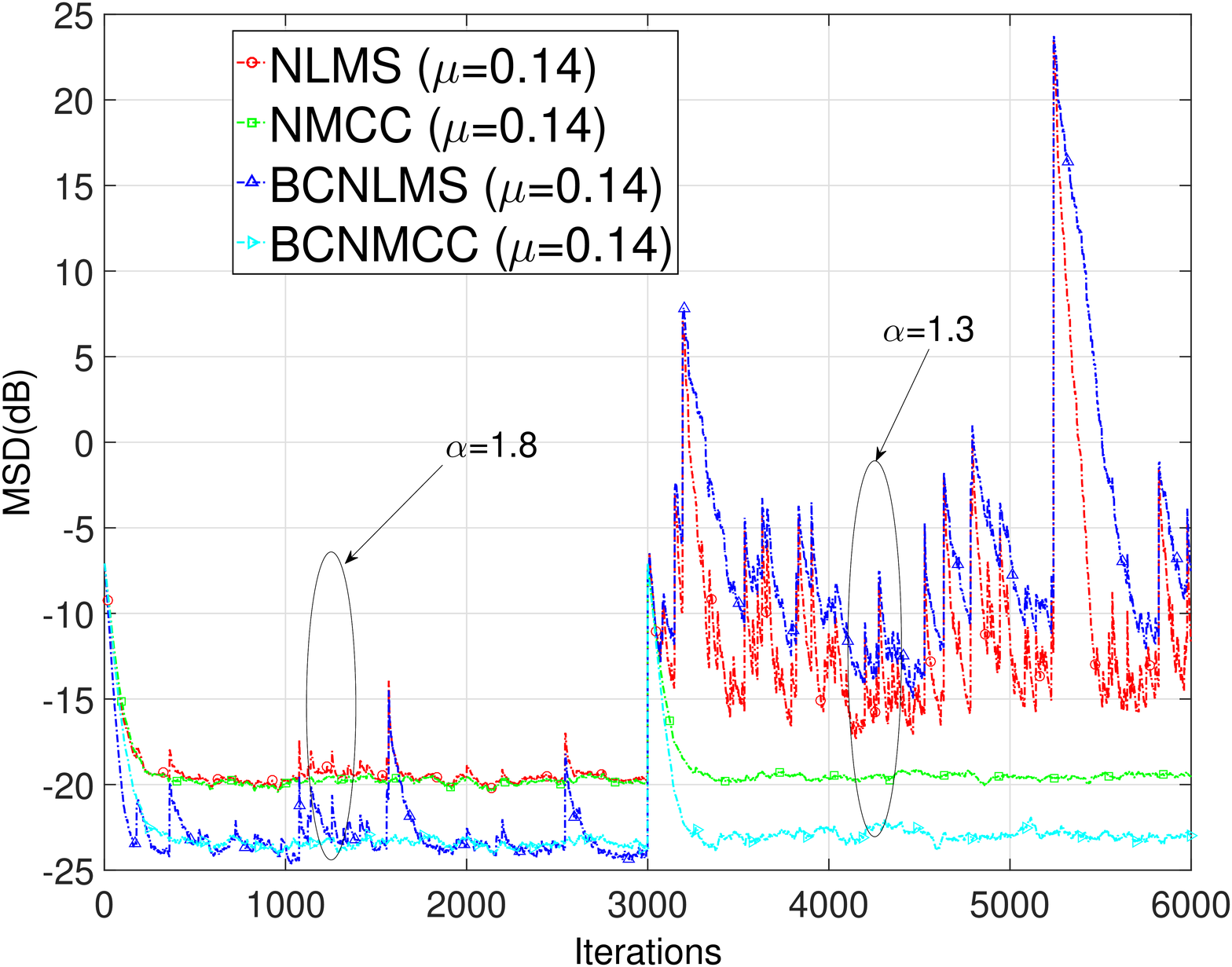}
\caption{Convergence curves of all algorithms under different $\alpha$ (1.8, 1.3).}
\label{fig_sim}
\end{figure}

In the second example, we only investigate the performance of the MCC and the BCNMCC with two stages. The step sizes are denoted as ${{\mu }_{1}}$ and ${{\mu }_{2}}$ in the first and second stage respectively. We maintain the same initial convergence speed and the same steady-state accuracy. In Fig.2, one can see that the BCNMCC has better steady-state performance than MCC in the first stage, while the convergence speed outperforms MCC at the second stage. This result demonstrates again that the proposed BCNMCC has better performance to the traditional MCC algorithm with noisy input in an impulsive noise environment.
\begin{figure}[!t]
\centering
\includegraphics[height=7cm,width=9.5cm]{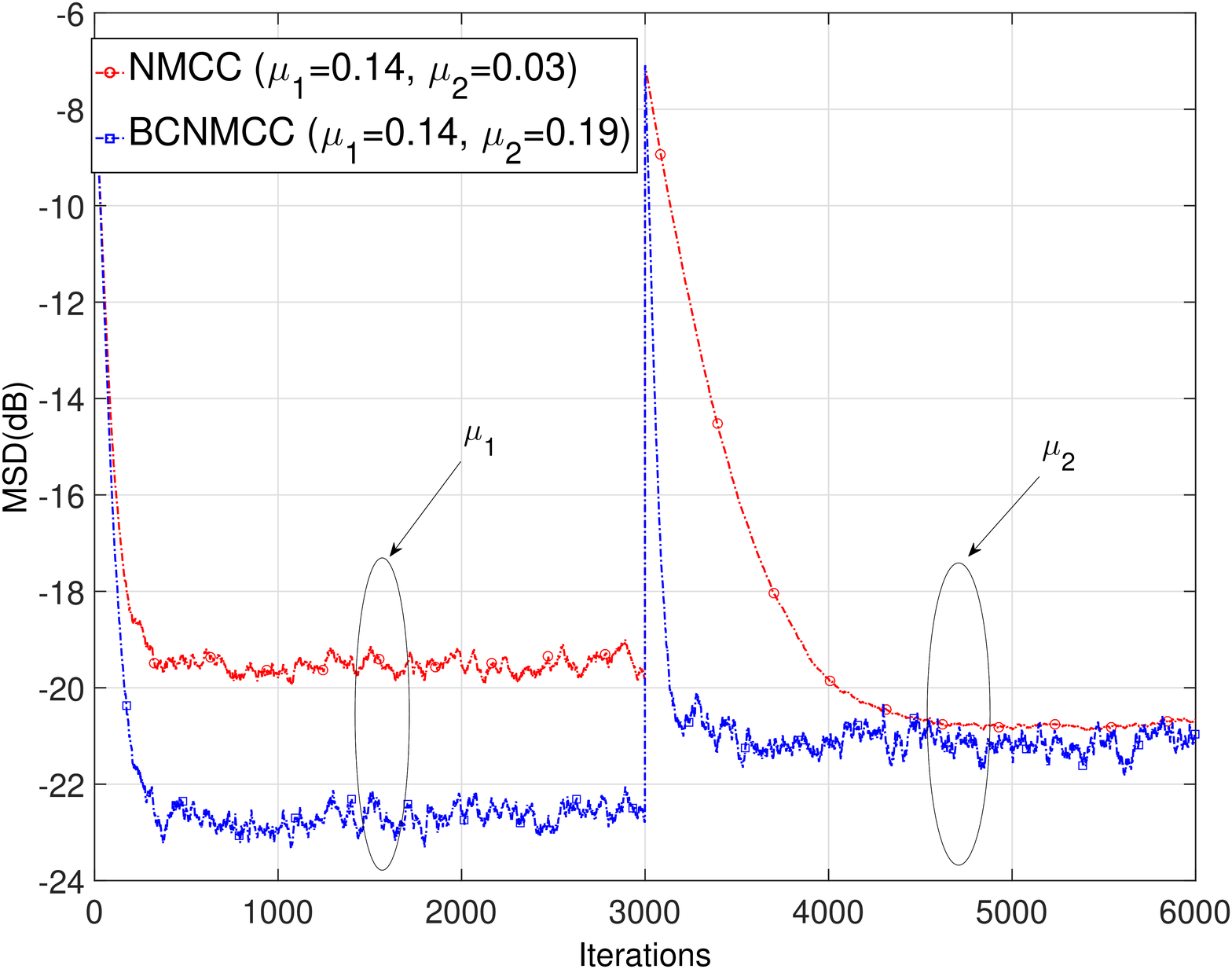}
\caption{Convergence curves of MCC and BCNMCC subject to the similar initial convergence speed and the similar steady-state accuracy respectively.}
\label{fig_sim}
\end{figure}

In the third example, we conduct simulations with different kernel size ( $\sigma $ = 3, 4, 5, 6 and 7) to evaluate the performance of the proposed method. Other parameters are set the same as the second example. The last 200 iterations results at final stead-state are averaged to compute the steady-state MSD (ssMSD), which is illustrated in Fig.3. It is obvious that the proposed algorithm outperforms traditional MCC algorithm under different $\sigma $.
\begin{figure}[!t]
\centering
\includegraphics[height=7cm,width=9.5cm]{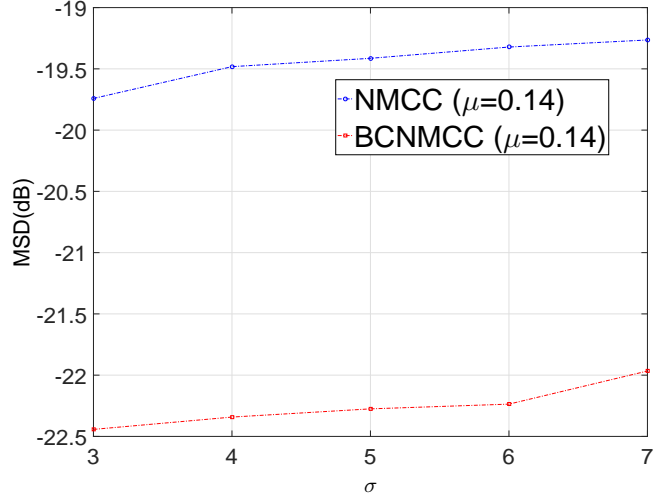}
\caption{Steady-state MSD with different values of $\sigma$.}
\label{fig_sim}
\end{figure}
\begin{figure}[!t]
\centering
\includegraphics[height=7cm,width=9.5cm]{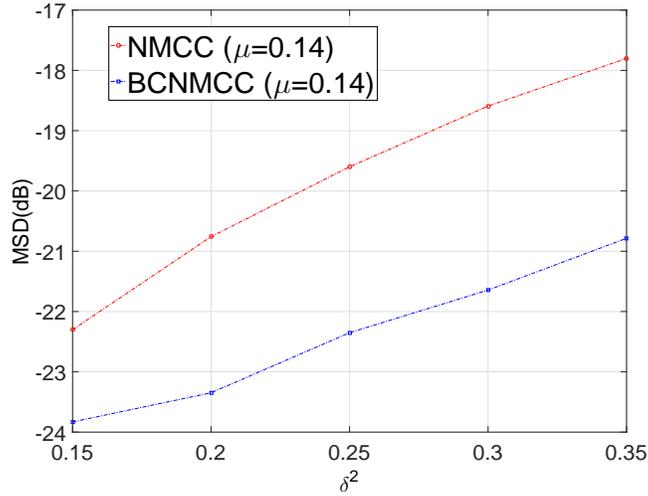}
\caption{Steady-state MSD results versus different variances of the input noise.}
\label{fig_sim}
\end{figure}

In the final example, we examine the performance of the proposed BCNMCC algorithm under different variances (0.15, 0.2, 0.25, 0.3, and 0.35) of the input noise. The ssMSD results are shown in Fig.4. Although the ssMSD results become worse with the variances increasing, we can observe that the performance of the BCNMCC algorithm is better than MCC algorithm under different variances, which is exactly what we expect.

\section{Conclusion}
The MCC based adaptive filter algorithms can offer better performance in impulsive output noise environments compared to many other methods. However, the input noise often exists in a real situation, which can significantly damage the performance of the adaptive filters. In order to address this problem, we combine the unbiasedness criterion (which has been successfully employed in several algorithms to address the bias caused by the input noise) and MCC to develop a novel robust bias-compensated NMCC algorithm, which can reduce the bad impact of the input noise, while maintaining the robustness of the NMCC with respect to impulsive output noise. Simulation results demonstrate that the proposed algorithm performs very well for SI problems with noisy input in an impulsive output noise environment.
\section*{Acknowledgements}
This work was supported in part by 973 Program (No. 2015CB351703), National Natural Science Foundation of China (No. 91648208, No. U1613219), the Natural Science Basic Research Plan in Shaanxi Province of China (No. 2017JM6033) and the Scientific Research Program Funded by Shaanxi Provincial Education Department (No.17JK0550).

\end{document}